\pdfoutput=1
\documentclass[11pt,a4paper]{article}
\usepackage[hyperref]{naaclhlt2018}

\usepackage{times}
\usepackage{url}
\usepackage{latexsym}
\usepackage{comment}

\usepackage[british]{babel}
\usepackage{graphicx}
\usepackage{colortbl}
\usepackage{hhline}
\usepackage{url}
\usepackage{latexsym}
\usepackage{amssymb}
\usepackage{amsmath}
\usepackage{enumerate}
\usepackage{algorithm}
\usepackage{algorithmicx}
\usepackage[noend]{algpseudocode}
\usepackage{booktabs}
\usepackage{proof}
\usepackage{paralist}
\usepackage{tikz-dependency}
\usepackage{tikz} 
\usepackage{graphicx, subfigure}
\usepackage{epstopdf}
\usepackage[latin1]{inputenc}

\newcommand{\eat}[1]{}

\makeatletter
\useshorthands{"}%
\defineshorthand{"-}{\nobreak-\bbl@allowhyphens}
\makeatother

\newcommand{\transname}[1]{\ensuremath{\mathsf{#1}}}

\newcommand{\stacktop}{{\mid}}

\mathchardef\mhyphen="2D 

\newcommand{\la}{\transname{Left\mhyphen Arc}}
\newcommand{\ra}{\transname{Right\mhyphen Arc}}

\newcommand{\sh}{\transname{Shift}}
\newcommand{\re}{\transname{Reduce}}
\newcommand{\swap}{\transname{Swap}}
\newcommand{\sw}{\transname{Switch}}

\newcommand{\squishlist}{ 
   \begin{list}{$\bullet$}
    { \setlength{\itemsep}{0pt}      \setlength{\parsep}{3pt} 
      \setlength{\topsep}{3pt}       \setlength{\partopsep}{0pt}
      \setlength{\leftmargin}{1.5em} \setlength{\labelwidth}{1em}
      \setlength{\labelsep}{0.5em} } }

\newcommand{\squishend}{
    \end{list}  } 

\aclfinalcopy 
\def\aclpaperid{744} 

\title{A Dynamic Oracle for Linear-Time 2-Planar Dependency Parsing}
  
\author{Daniel Fern\'{a}ndez-Gonz\'{a}lez \and Carlos G\'{o}mez-Rodr\'{i}guez\\
	Universidade da Coru\~{n}a\\
	FASTPARSE Lab, LyS Research Group, Departamento de Computaci\'{o}n \\
	Campus de Elvi\~{n}a, s/n, 15071 A Coru\~{n}a, Spain \\
  {\tt d.fgonzalez@udc.es}, {\tt carlos.gomez@udc.es}\\}

\date{}

\begin{document}
\maketitle
\begin{abstract}
We propose an efficient dynamic oracle for training the 2-Planar transition-based parser, 
a linear-time parser with over 99\% coverage on non-projective syntactic corpora. 
This novel approach outperforms the static training strategy in the vast majority of languages tested and scored better on most datasets than the arc-hybrid parser enhanced with the $\swap$ transition, which can handle unrestricted non-projectivity.
\end{abstract}

\section{Introduction}
Linear-time greedy transition-based parsers such as \textit{arc-eager}, \textit{arc-standard} and \textit{arc-hybrid} \cite{Nivre2003,nivre04acl,archybrid} are widely used for dependency parsing due to their efficiency and performance, but they cannot deal with non-projective syntax.
To address this, various extensions
have been proposed, involving new transitions \cite{attardi06,nivre09acl,buffertrans}, data structures \cite{twoplanar,PitlerMcDonald2015} or pre and postprocessing \cite{NivNil05}. 
Among these extensions, the 2-Planar parser \cite{twoplanar} has attractive properties, as it (1) keeps the original worst-case linear time, (2) has close to full coverage of non-projective phenomena, and (3) needs no pre- or post-processing.

Dynamic oracles \cite{goldberg2012dynamic} are known to improve the accuracy of greedy parsers by enabling 
more robust training, by exploring configurations beyond the gold path.
While dynamic oracles have been defined for many transition-based algorithms
\cite{goldberg2013training,goldberg2014tabular,Gomez14dynamic,dyncovington,delhoneux17arc},
none is available so far for the 2-Planar system.
The lack of the arc-decomposability property, which can be used to derive dynamic oracles for parsers that have it, makes the obtention of one non-trivial.

To fill this gap, we define an efficient dynamic oracle for the 2-Planar transition-based parser, using similar loss calculation techniques as described in \cite{dyncovington} for the non-arc-decomposable Covington parser \cite{covington01fundamental}.
Training the 2-Planar parser with this novel strategy achieves accuracy gains in the vast majority of datasets tested. In addition, we empirically compare our novel approach to the most similar existing alternative:\footnote{Although the two-register parser by \newcite{PitlerMcDonald2015} is even closer to ours in terms of the transition system (being based on arc-eager and running in linear time), no dynamic oracle is known for it, to the best of our knowledge.} the arc-hybrid parser with a swap transition trained with a static-dynamic oracle, recently introduced by \newcite{delhoneux17arc}; which can handle unrestricted non-projective dependencies in $O(n^2)$ worst-case time in theory, but expected linear time in practice \cite{nivre09acl}. Our approach outperforms this swap-based system on average over a standard set of dependency treebanks.

\section{The 2-Planar parser}
We briefly sketch the 2-Planar transition system, which was defined by \newcite{twoplanar,GomNivCL2013} under the transition-based parsing framework \cite{Nivre2008} and is based on the arc-eager algorithm \cite{Nivre2003}, keeping its linear time complexity. 
It works by building, in a single pass, two non-crossing graphs (called \emph{planes}) whose union provides a dependency parse in the set of 2-planar (or pagenumber-2) graphs, which is known to cover over 99\% of parses in a large number of real treebanks \cite{twoplanar,GomCL2016}.

Parser configurations have the form {$c=\langle {\Sigma_1} , {\Sigma_2} , {B} , {A} \rangle$}, where $\Sigma_1$ and $\Sigma_2$ are, respectively, the \textit{active stack} and \textit{inactive stack} 
and are tied to each of the planes or pages of the output graph, $B$ is a \textit{buffer} of unread words, and $A$ the set of dependency arcs built so far. For an input string $w_1 \cdots w_n$, the parser starts at configuration $c_s(w_1 \ldots w_n) = \langle [\ ],[\ ],[w_1 \ldots w_n],\emptyset \rangle$,
applying transitions until
a terminal configuration $\langle \Sigma_1, \Sigma_2 , [\ ] , A \rangle$ is reached, and $A$ yields the output.

Figure~\ref{fig:transitions} shows the 
parser's transitions.
The $\sh$ transition pops the first (leftmost) word in the buffer, and pushes it to both stacks; the $\re$ transition pops the top word from the active stack, implying that we have added all arcs to or from it on the plane tied to that stack; and the $\la$ and $\ra$ transitions create a leftward or rightward dependency arc connecting the first word in the buffer to the top of the active stack. Finally, the $\sw$ transition makes the active stack inactive and vice versa, changing the plane the parser is working with. 
Transitions that violate the single-head or acyclicity constraints are disallowed, so that the output is a forest.
Finally, to guarantee the termination of the parsing process, two consecutive $\sw$ transitions are not allowed.

\begin{figure*}
\begin{tabbing}
\hspace{2cm}\=\hspace{2.4cm}\= \kill
\> \sh: \> $\langle {\Sigma_1}, {\Sigma_2}, {w_i \stacktop B} , {A} \rangle
\Rightarrow \langle {\Sigma_1 \stacktop w_i} , {\Sigma_2 \stacktop w_i}, B , A \rangle$ { }\\[1mm]
\> \re: \> $\langle {\Sigma_1 \stacktop w_i}, {\Sigma_2}, {B} , {A} \rangle
\Rightarrow \langle {\Sigma_1} , {\Sigma_2}, B , A \rangle$ { }\\[1mm]
\> \la: \> $\langle {\Sigma_1 \stacktop w_i} , {\Sigma_2}, {w_j \stacktop B} ,
A \rangle \Rightarrow
\langle {\Sigma_1 \stacktop w_i}, {\Sigma_2}, {w_j \stacktop B} , A \cup \{ w_j \rightarrow w_i \} \rangle$\\
\> \> \small{only if $\nexists w_k \mid w_k \rightarrow w_i \in A$ (single-head) and $w_i  \rightarrow^\ast w_j \not\in A$ (acyclicity).}\\[1mm]
\> \ra: \> $\langle {\Sigma_1 \stacktop w_i} , {\Sigma_2}, {w_j \stacktop B} ,
A \rangle \Rightarrow
\langle {\Sigma_1 \stacktop w_i}, {\Sigma_2}, {w_j \stacktop B} , A \cup \{ w_i \rightarrow w_j \} \rangle$\\
\> \> \small{only if $\nexists w_k \mid w_k \rightarrow w_j \in A$ (single-head) and $w_j  \rightarrow^\ast w_i \not\in A$ (acyclicity).}\\[1mm]
\> \sw: \> $\langle {\Sigma_1}, {\Sigma_2}, B , A \rangle
\Rightarrow \langle \Sigma_2, {\Sigma_1  }, B , A \rangle$ { } \\
\end{tabbing}

\caption{Transitions of the 2-Planar dependency parser. The notation $w_i \rightarrow^\ast w_j \in A$ means that there is a (possibly empty) directed path from $w_i$ to $w_j$ in $A$.}
\label{fig:transitions}
\end{figure*}

\section{A dynamic oracle}
We now define an efficient dynamic oracle to train the 2-Planar algorithm, which operates under the assumption of a fixed assignment of arcs to planes.

Following \newcite{goldberg2013training}, if the Hamming loss ($\mathcal{L}$) between trees $t$ and $t_G$ is the amount of words with a different head in $t$ and $t_G$, then implementing a dynamic oracle reduces to defining a loss function $\ell(c)$ which, given a parser configuration $c$ and a gold tree $t_G$, computes the minimum loss between $t_G$ and a tree reachable from $c$.\footnote{We say that an arc set $X$ is \textit{reachable} from configuration $c$, and we write $c \rightsquigarrow X$, if there is some (possibly empty) path of transitions from $c$ to some configuration $c'=\langle {\Sigma_1} , {\Sigma_2} , {B} , {A'} \rangle$, with $X \subseteq A'$.} We call this the minimum loss of configuration $c$, 
$\ell(c) = \min_{t | c \rightsquigarrow t} \mathcal{L}(t,t_G)$. 
\noindent A correct dynamic oracle will return the set of transitions $\tau$ that do not increase this loss (i.e., $\ell( \tau(c) ) - \ell(c) = 0 $), thus leading to the best parse reachable from $c$.

For parsers that are \textit{arc-decomposable}\footnote{i.e., if every individual arc of $X$ is reachable from a given configuration $c$, the set $X$ as a whole is reachable from $c$.}, $\ell(c)$ can be obtained by 
counting 
gold arcs that are \emph{not} individually reachable from $c$, which is trivial in most parsers. Unfortunately, the 2-Planar parser is non-arc-decomposable. To show this, it suffices to consider any configuration where an incorrect arc created in $A$ forms a cycle together with a set of 
otherwise reachable gold arcs,
just as in the proof of non-arc-decomposability for Covington provided by \newcite{dyncovington}. In fact, the same counterexample provided there also works for this parser. 

Note, however, that non-arc-decomposability in the 2-Planar parser not only comes from cycles (as in Covington) but also from  
situations where, due to 
a poor 
assignment of planes to already-built arcs, no possible plane assignment allows building a set of pending gold arcs.
Thus, the loss calculation technique of the Covington dynamic oracle is not directly applicable to the 2-Planar parser. 

However, if we statically choose a canonical plane assignment and we calculate loss with respect to that  assignment (i.e., creating a correct arc in the non-canonical plane incurs loss), then the Covington technique, based on counting individually unreachable arcs and then correcting for the presence of cycles, works for the 2-Planar parser. This is the idea of our dynamic oracle, which therefore is a correct dynamic oracle only with respect to a preset criterion for plane assignment, and not for all the possible plane assignments that would produce the gold dependency structure.

In particular, given a 2-planar gold dependency tree whose set of arcs is $t_G$, we need to divide it into two gold arc sets $t_G^1$ and $t_G^2$, associated with each plane.\footnote{In practice, for gold parses that are not 2-planar, some arcs will need to be discarded, so that $t_G^1 \cup t_G^2$ will be a 2-planar subset of $t_G$. Note that in this case, our oracle is correct with respect to this 2-planar subset, but it does not guarantee minimum loss with respect to the original non-2-planar graph (in the same way as existing projective dynamic oracles do not guarantee it with respect to non-projective trees).} In this paper, we take as canonical the division provided by the static oracle of \newcite{twoplanar}, which prefers to build arcs in the active plane to minimize the number of {\sw} transitions needed.\footnote{Our dynamic oracle can work with any plane assignment criterion. We chose this one for simplicity, for direct comparability to the existing static oracle, and because it has been shown to be learnable in practice.}

Once the plane assignment is set, 
we can associate individually unreachable arcs to a plane.
Then, we can 
calculate configuration loss as:
\begin{multline*}
\ell(c) = |{\mathcal{U}_1(c,t_G^1)} \cup {\mathcal{U}_2(c,t_G^2)}| \\ 
+ n_c(A \cup ({\mathcal{I}_1(c,t_G^1)} \cup {\mathcal{I}_2(c,t_G^2)}))
\end{multline*}
\noindent where for $i \in \{1,2\}$, each set $\mathcal{I}_i(c,t_G^i) = \{ x \rightarrow y \in t_G^i \mid c \rightsquigarrow (x \rightarrow y) \}$ is the set of 
\textit{individually reachable arcs} of $t_G^i$  from configuration $c$; $\mathcal{U}_i(c,t_G^i)$ is the set of \textit{individually unreachable arcs} of $t_G^i$ from $c$, defined as $ t_G^i \setminus \mathcal{I}_i(c,t_G^i)$; and $n_c(G)$ denotes the number of cycles in a graph $G$.

To compute the sets of individually unreachable arcs $\mathcal{U}_i(c,t_G^i)$ from a configuration $c =\langle {\Sigma_1} , {\Sigma_2}, {B} , A \rangle$, we examine gold arcs. A gold arc $x \rightarrow y \in t_G^i$ will be in $\mathcal{U}_i(c,t_G^i)$ if it is not in $A \cap t_G^i$ (the set of already-built arcs from the plane of interest), and at least one of the following holds:

\begin{itemize}
\item  $\min(x,y) \not\in \Sigma_i  \cup B \vee \max(x,y) \not\in B$, (i.e., $\min(x,y)$ must be in plane $i$'s stack or in the buffer, and $\max(x,y)$ must be in the buffer so that the arc $x \rightarrow y$ can still be built),
\item there is some $z \neq 0, z \neq x$ such that $z \rightarrow y \in A$, (i.e., we cannot create $x \rightarrow y$ because it would violate the single-head constraint),
\item $x$ and $y$ are on the same weakly connected component of $A$ (i.e., we cannot create $x \rightarrow y$ due to the acyclicity constraint).
\item $x \rightarrow y \in A \cap t_G^{3-i}$ (i.e., the arc was already erroneously created in the other plane and, therefore, is unreachable in plane $i$).
\end{itemize}

Once we have $\mathcal{U}_i(c,t_G^i)$ for each of the two planes, ${\mathcal{I}_i(c,t_G^i)}$ can be obtained as $t_G^i \setminus \mathcal{U}_i(c,t_G^i)$.
Finally, since the graph $A \cup {\mathcal{I}_1(c,t_G^1)} \cup {\mathcal{I}_2(c,t_G^2)}$ has in-degree 1, the algorithm by \newcite{Tarjan72} can be used to implement the function $n_c$ to count its cycles in $O(n)$ time. For this reason, the full loss calculation runs in linear time as well.\footnote{The check for acyclicity using weakly connected components has no impact on the complexity: when weakly connected components are represented using path compression and union by rank, the relevant operations run in amortized inverse Ackermann time, meaning that they behave as constant time for all practical purposes, like in \cite{GomNivCL2013}}

Given a plane assignment, $\ell(c)$ is an exact expression of the loss of a configuration of the 2-Planar parser as expressed in Figure \ref{fig:transitions}, without the control constraint that forbids two consecutive $\sw$ transitions. This can be proven using the same reasoning as for the Covington loss expression of \cite{dyncovington}. Thus, the computation of $\ell(c)$ provides a complete and correct dynamic oracle for this parser under a given plane assignment, by directly evaluating $\ell(\tau(c))-\ell(c)$ for each transition $\tau$. However, to make the oracle correct for the practical version, where consecutive $\sw$ transitions are disallowed, we need to modify the cost calculation for the $\sw$ transition.

In particular, applying a $\sw$ transition does not affect the loss, so
$\ell(\sw(c))-\ell(c)$ is always $0$. Indeed, if $\sw$ transitions are always allowed, their cost is zero because they can always be undone and thus never affect the reachability of any arcs. However, when consecutive $\sw$ transitions are banned to ensure parser termination, choosing to $\sw$ can have consequences as, in the resulting  configuration, the parser will be forced to take one of the other four transitions, which may lead to suboptimal outcomes  compared to not having switched. 

To address this, we compute the cost of $\sw$ transitions instead as $min(\{\ell( \tau(\sw(c)))-\ell(c) | \tau \neq \sw \})$, i.e., 
the minimum number of gold arcs missed after being forced to apply one of the other four transitions after $\sw$ (if this cost is $0$, then switching stacks is an optimal choice). Adding this modification makes the dynamic oracle correct for the practical version of the parser that disallows consecutive $\sw$ transitions.

\paragraph{Regularization}

While the above dynamic oracle is theoretically correct, we noticed experimentally that the $\sw$ transition tends to switch stacks very frequently during training, due to exploration. This leads the parser to learn unnecessarily long and complex transition sequences that change planes more than needed, harming accuracy.

To avoid this, we add a regularization term to $\ell(c)$ representing the transition sequence length from $c$ to its minimum-loss reachable tree(s), to discourage unnecessarily long sequences. This amounts to penalizing the $\sw$ transition if there is any zero-cost transition available in the active plane and changing planes will delay its application. Thus, arcs assigned to the currently active plane will be built before switching if possible, enforcing a global arc creation order. This is similar to the prioritization of monotonic paths in \cite[\S 6]{honnibal13}, as they also penalize unneeded actions that will need to be undone later.

\section{Experiments}
\subsection{Data and Evaluation}
We conduct our experiments on the commonly-used non-projective benchmark compounded of nine datasets from the CoNLL-X shared task \cite{buchholz06} and all datasets from the CoNLL-XI shared task \cite{conll2007}.\footnote{We use  for evaluation the latest version for each language, i.e., if a language appeared in both CoNLL-X and CoNLL-XI, we use the CoNLL-XI dataset.} We also use the Stanford Dependencies \cite{deMarneffe2008} conversion (using the Stanford parser v3.3.0)\footnote{\url{https://nlp.stanford.edu/software/lex-parser.shtml}} of the WSJ Penn Treebank (PTB-SD) \cite{marcus93} with standard splits. Labelled and Unlabelled Attachment Scores (LAS and UAS) are computed
including punctuation for all datasets except for the PTB where, following common practice, the punctuation is excluded. We train our system for 15 iterations and choose the best model according to 
development set accuracy.
Statistical significance is calculated 
using
a  paired  test  with  10,000 bootstrap  samples. 

\subsection{Model}
We implement both the static oracle and the dynamic oracle with aggressive exploration for the 2-Planar parser under the neural network architecture proposed by \newcite{Kiperwasser2016}.
We also add the static-dynamic arc-hybrid parser with $\swap$ transition \cite{delhoneux17arc}, implemented under the same framework to perform a fair comparison. 

The neural network architecture used in this paper is taken from \newcite{Kiperwasser2016}. We use the same BiLSTM-based featurization method that concatenates the representations of the top 3 words on the active stack and the leftmost word in the buffer for the arc-hybrid and 2-Planar algorithms, and we add the top 2 words on the inactive stack for the latter. Following \newcite{Kiperwasser2016}, we also include  the BiLSTM vectors of the  rightmost and leftmost modifiers of words from the  stacks, as well as the leftmost modifier of the first word in the buffer. We initialize word embeddings with 100-dimensional GloVe vectors \cite{glove} for English and use 300-dimensional Facebook vectors \cite{facebookemb} for other languages. The other parameters of the neural network keep the same values as in \cite{Kiperwasser2016}.  

\begin{table}
\begin{center}
\centering
\begin{tabular}{@{\hskip 0pt}l|cc|cc@{\hskip 0pt}}
& \multicolumn{2}{c|}{2-Planar}
& \multicolumn{2}{c}{2-Planar}
\\
& \multicolumn{2}{c|}{static}
& \multicolumn{2}{c}{dynamic}
\\
Language
& UAS & LAS
& UAS & LAS
\\
\hline
Arabic & \textbf{83.20} & \textbf{73.48} & 82.96 &  73.24 \\ 
Basque &  77.61 & 69.94 & \textbf{78.11} & \textbf{70.11} \\ 
Catalan & 92.50 & 87.92 & \textbf{93.70$^{*}$} & \textbf{88.48$^*$} \\ 
Chinese & 85.95 & 80.97 & \textbf{87.08}$^*$ & \textbf{81.73}$^*$ \\  
Czech & 84.67 & 78.56 & \textbf{85.29} & \textbf{79.40}    \\ 
English & 89.57 & 88.69  & \textbf{90.87}$^*$  &  \textbf{90.03}$^*$ \\  
Greek & 81.93 & 74.23  & \textbf{82.06} & \textbf{74.79}  \\ 
Hungarian & 81.50 & 75.94 & \textbf{82.48$^*$} & \textbf{76.97$^*$} \\ 
Italian & 86.85 & 82.36 & \textbf{87.24} & \textbf{82.38} \\ 
Turkish & \textbf{81.68} & 73.59 & 81.48 & \textbf{73.61} \\ 
\hline
Bulgarian & 93.14 & 89.74 & \textbf{93.23} & \textbf{89.97} \\ 
Danish & 88.31 & 84.23 & \textbf{88.57} & \textbf{84.76}  \\ 
Dutch & 85.51 & 81.63 & \textbf{86.50$^{*}$} & \textbf{82.70$^{*}$} \\ 
German &  \textbf{90.80} & \textbf{88.71} & 90.71 & 88.51 \\
Japanese & 92.51 & 90.56 & \textbf{93.19}$^*$ & \textbf{90.65}$^*$  \\ 
Portugue. & 88.68 & 85.12 & \textbf{89.02$^*$} & \textbf{85.92$^*$} \\ 
Slovene & 78.67 & 70.49 &  \textbf{79.30} & \textbf{70.86} \\ 
Spanish & \textbf{83.63}$^*$ & \textbf{79.79}$^*$ & 82.42 & 78.68 \\ 
Swedish &  \textbf{89.92} & \textbf{85.48} & 89.83 & 85.40 \\ 
\hline
PTB-SD & 93.59  & 91.60 & \textbf{93.96}$^*$ & \textbf{92.06}$^*$ \\
\hline 
Average & 86.51  & 81.65 & \textbf{86.90} & \textbf{82.01}  \\
\hline
\multicolumn{5}{c}{}\\
\end{tabular}
\centering
\caption{Parsing accuracy of the 2-Planar parser trained with static and dynamic oracles on CoNLL-XI (first block), CoNLL-X (second block) and PTB-SD (third block) datasets. Best results for each language are shown in boldface. Statistically significant improvements ($\alpha = .05$) are marked with $^*$. }
\label{tab:results}
\end{center}
\end{table}

\begin{table}
\begin{center}
\centering
\begin{tabular}{@{\hskip 0pt}l|cc|cc@{\hskip 0pt}}
& \multicolumn{2}{c|}{2-Planar}
& \multicolumn{2}{c}{AHybrid$_{\swap}$}
\\
& \multicolumn{2}{c|}{dynamic}
& \multicolumn{2}{c}{\small{static-dynamic}}
\\
Language
& UAS & LAS
& UAS & LAS
\\
\hline
Arabic & \textbf{82.96}$^*$ &  \textbf{73.24}$^*$ & 80.74 & 70.69 \\ 
Basque & \textbf{78.11$^*$} & \textbf{70.11$^*$} &  75.60 & 68.70  \\ 
Catalan & \textbf{93.70$^{*}$} & \textbf{88.48} & 93.12 & 88.06  \\ 
Chinese & 87.08 & 81.73  & \textbf{87.31} & \textbf{82.02}  \\  
Czech & 85.29 & 79.40   & \textbf{85.71$^*$} & \textbf{80.08$^*$}  \\ 
English & 90.87  &  90.03 & \textbf{91.37} & \textbf{90.37}  \\  
Greek & 82.06 & 74.79  & \textbf{83.72$^*$} & \textbf{76.33$^*$}  \\ 
Hungarian & \textbf{82.48} & \textbf{76.97} &  82.20 & 76.88  \\ 
Italian & \textbf{87.24$^*$} & \textbf{82.38$^*$}  & 86.24 & 81.48  \\ 
Turkish & \textbf{81.48}$^*$ & \textbf{73.61$^*$} &   77.44 & 69.22  \\ 
\hline
Bulgarian & \textbf{93.23} & \textbf{89.97} &  93.17 & 89.62  \\ 
Danish & 88.57 & \textbf{84.76} &  \textbf{88.65} & 84.60  \\ 
Dutch & \textbf{86.50$^{*}$} & \textbf{82.70$^{*}$} &  84.24 & 81.04  \\ 
German & \textbf{90.71} & \textbf{88.51} &  90.60 & 88.39   \\ 
Japanese & 93.19 & 90.65  & \textbf{93.40} & \textbf{91.47$^*$}  \\ 
Portugue. & \textbf{89.02} & \textbf{85.92} &  88.78 & 85.48  \\ 
Slovene & \textbf{79.30$^*$} & \textbf{70.86}  & 77.68 & 70.61  \\ 
Spanish & 82.42 & 78.68 &  \textbf{83.98$^*$} & \textbf{80.15$^*$}  \\ 
Swedish & 89.83 & \textbf{85.40} &  \textbf{89.92} & 85.18  \\ 
\hline
PTB-SD & \textbf{93.96} & \textbf{92.06} & 93.83  &  91.93\\
\hline 
Average &\textbf{86.90} & \textbf{82.01} &  86.39 & 81.62  \\
\hline
\multicolumn{5}{c}{}\\
\end{tabular}
\centering
\caption{Parsing accuracy of the 2-Planar parser trained with the dynamic oracle and the arc-hybrid parser with the $\swap$ transition trained with a static-dynamic oracle on CoNLL-XI (first block), CoNLL-X (second block) and PTB-SD (third block) datasets. Best results for each language are in boldface. Statistically significant improvements ($\alpha = .05$) are marked with $^*$. }
\label{tab:results2}
\end{center}
\end{table}

\subsection{Results}
Table~\ref{tab:results} shows that the 2-Planar parser trained with a dynamic oracle outperforms the static training strategy in terms of UAS in 15 out of 20 languages, with 8 of these improvements  statistically significant ($\alpha = .05$), and one statistically significant decrease. When comparing with the enhanced arc-hybrid system in Table~\ref{tab:results2}, our approach provides a better UAS in 12 out of 20 datasets tested, achieving statistically significant ($\alpha = .05$) gains in accuracy on 7 of them, and significant losses on 3 of them.

We could not find a clear pattern to explain why the 2-Planar algorithm outperforms arc-hybrid plus $\swap$ in some languages and vice versa.
The latter seems to work better on treebanks with less non-projectivity such as the English, Chinese and Japanese datasets, and worse on those with higher amounts like Turkish, Dutch or Basque.
However, some cases like Czech or Catalan go against this trend.
From \cite{twoplanar}, we also know that the Dutch and German treebanks have a relatively high proportion of non-2-planar trees, but the 2-Planar parser seems to be a better option on them than the extended arc-hybrid system that can handle unrestricted non-projectivity.
The reasons, beyond the scope of this research, might be related to different dependency length distributions or non-projective topologies.

We noticed that, in general, the 2-Planar parser has higher precision 
on non-projective arcs and the enhanced arc-hybrid parser has a better recall.

\section{Conclusion}
We present an efficient dynamic oracle to train the 2-Planar transition-based parser, which is correct with respect to a given plane assignment, and results in notable gains in accuracy. The parser trained with this dynamic oracle performs better on average than 
an expected linear-time parser supporting unrestricted non-projectivity.

\section*{Acknowledgments}

This work has received funding from the European
Research Council (ERC), under the European
Union's Horizon 2020 research and innovation
programme (FASTPARSE, grant agreement No
714150), from the TELEPARES-UDC project
(FFI2014-51978-C2-2-R) and the ANSWER-ASAP project (TIN2017-85160-C2-1-R) from MINECO, and from Xunta de Galicia (ED431B 2017/01).

\bibliography{main,twoplanaracl,bibliography}
\bibliographystyle{acl_natbib}

\clearpage
\appendix

\end{document}